%% file: main.tex
\documentclass[journal]{IEEEtran}
\usepackage{cite}
\usepackage{amsmath,amssymb,amsfonts}
\usepackage{algorithmic}
\usepackage{graphicx}
\usepackage{textcomp}

\usepackage[utf8]{inputenc}
\usepackage{graphicx}
\usepackage{comment}
\usepackage{color}
\usepackage{multirow}
\usepackage{booktabs}

\usepackage{array}

\newcolumntype{Y}{>{\centering\arraybackslash}m{1cm}}
\usepackage{footmisc}
\usepackage{siunitx,booktabs,etoolbox}
\usepackage[colorlinks=true,citecolor=blue,urlcolor=black]{hyperref}
\usepackage[misc]{ifsym}
\usepackage{algorithm}

\providecommand{\mb}[1]{\mathbf{#1}}

\providecommand{\mcl}[1]{\mathcal{#1}} 
\providecommand{\redc}[1]{\textcolor{red}{#1}} 
\providecommand{\bluec}[1]{\textcolor{blue}{#1}} 

\providecommand{\mbx}{\mb{x}}
\providecommand{\mby}{\mb{y}}

\def\BibTeX{{\rm B\kern-.05em{\sc i\kern-.025em b}\kern-.08em
    T\kern-.1667em\lower.7ex\hbox{E}\kern-.125emX}}
\begin{document}
\title{Causality-inspired Single-source Domain Generalization for Medical Image Segmentation}
\author{Cheng Ouyang, Chen Chen, Surui Li, Zeju Li, Chen Qin, Wenjia Bai and Daniel Rueckert
\thanks{This work is support by EPSRC Programme Grant EP/P001009/1. (Corresponding author: Cheng Ouyang, E-mail: c.ouyang@imperial.ac.uk)}
\thanks{Cheng Ouyang, Chen Chen, Surui Li, Zeju Li, Wenjia Bai and Daniel Rueckert are with the Department of Computing, Imperial College London, UK; Chen Qin is with the Institute for Digital Communications, School of Engineering, University of Edinburgh, UK; Wenjia Bai is also with the Department of Brain Sciences, Imperial College London, UK; Daniel Rueckert is also with the Institute for AI and Informatics in Medicine, Klinikum rechts der Isar, Technical University of Munich, Germany.}
}
\maketitle
\begin{abstract}
Deep learning models usually suffer from domain shift issues, where models trained on one source domain do not generalize well to other unseen domains. In this work, we investigate the single-source domain generalization problem: 
training a deep network that is robust to unseen domains, under the condition that training data is only available from one source domain, which is common in medical imaging applications. 
We tackle this problem in the context of cross-domain medical image segmentation. Under this scenario, domain shifts are mainly caused by different acquisition processes. 
We propose a simple causality-inspired data augmentation approach to expose a segmentation model to synthesized domain-shifted training examples. 
Specifically, 1) to make the deep model robust to discrepancies in image intensities and textures, we employ a family of randomly-weighted shallow networks. They augment training images using diverse appearance transformations. 2) Further we show that spurious correlations among objects in an image are detrimental to domain robustness. These correlations might be taken by the network as domain-specific clues for making predictions, and they may break on unseen domains. We remove these spurious correlations via causal intervention. This is achieved by resampling the appearances of potentially correlated objects independently. 
The proposed approach is validated on three cross-domain segmentation tasks: cross-modality (CT-MRI) abdominal image segmentation, cross-sequence (bSSFP-LGE) cardiac MRI segmentation, and cross-center prostate MRI segmentation. The proposed approach yields consistent performance gains compared with competitive methods when tested on unseen domains.
\end{abstract}

\input{1_introduction.tex}
\input{2_related_works.tex}

\input{3_method.tex}
\input{4_experiments.tex}

\input{5_discussion.tex}
\bibliographystyle{IEEEtran}
\bibliography{main.bbl}
\end{document}

%% file: 1_introduction.tex
\section{Introduction}
\label{sec: intro}
Deep learning based medical image segmentation approaches \cite{ronneberger2015u,shen2017deep,isensee2018nnu,zhang2020generalizing,zhou2018unet++,antonelli2021medical} usually achieve state-of-the-art performance when being trained and tested on datasets from a single \textit{domain}, \textit{i.e.} from identically distributed training and testing data. However, in practice, deep learning models perform less well when the testing data is drawn from a different distribution than the training data (\textit{i.e.} a different \textit{domain}) \cite{kamnitsas2017unsupervised,dou2018unsupervised}. The discrepancy between training and testing domains is termed as the \textit{domain shift} \cite{ben2010theory}. In medical image segmentation, the most notorious source of domain shift is the differences in image acquisition (imaging modalities, scanning protocols, or device manufacturers) \cite{glocker2019machine}. This type of domain shift is therefore termed as the \textit{acquisition shift} \cite{castro2020causality}. We argue that the performance deterioration under acquisition shift can be attributed to the following two mechanisms: the \textit{shifted domain-dependent features} and the \textit{shifted-correlation effect}.\\
\indent \textbf{Shifted domain-dependent features:} Domain-dependent features include \textit{intensities} and \textit{textures}, which constitute image \textit{appearance}. Deep networks are susceptible to shifts in intensity/texture \cite{dou2018unsupervised,kamnitsas2017unsupervised}. This is in contrast to human annotators: they can easily find correspondence of the same anatomical structure across domains \cite{dou2018unsupervised}, usually by looking at the \textit{shape} that is domain-invariant and intuitively \textit{causal} to human-defined segmentation masks, compared with intensity/texture. \\
\indent \textbf{Shifted-correlation effect:} Due to a \textit{confounder} (\textit{i.e.} a ``third'' variable that spuriously correlates two variable-of-interests) \cite{pearl1995causal,pearl2009causality}, objects in the background might be correlated but \textbf{not} \textit{causally} related to the object-of-interests \cite{zhang2020causal}. The network might take these objects in the background as clues for recognizing the object-of-interests \cite{zhang2020causal}. For example, in \cite{zech2018variable,geirhos2020shortcut}, a model that recognizes \texttt{pneumonia} in X-ray images is actually looking at the \texttt{hospital mark} in the background, which correlates with \texttt{pneumonia} due to the confounder: data selection bias. These correlations are often detrimental under domain shift. This is because decision rules based on these correlations may break in the shifted domain: the correlated objects in the background may disappear, or it may not co-shift in the same way as the object-of-interests. In the above example, the model fails on real-world images where \texttt{hospital mark} do not correlate to \texttt{pneumonia}.

To mitigate domain shift, previous attempts include \textit{unsupervised domain adaptation} (UDA) \cite{ganin2015unsupervised} and \textit{multi-source domain generalization} (MDG) approaches \cite{muandet2013domain}. Unfortunately, UDA or MDG may not always be practical: they rely on training data from the target domain or from multiple source domains, which are often unavailable due to cost or privacy concerns. UDA also requires expertise for fine-tuning on target data, incurring difficulties to its deployment in the real world.

A more practical setting is \textit{single-source domain generalization}: to train a deep learning model to be robust against domain shifts, using training data from only \emph{one} source domain. Since no examples of the target domain are available, we resort to bottom-up approaches that are built on the above causal analysis of acquisition shift. We aim to 1) steer the network towards shape information which is domain-invariant and is intuitively causal to segmentation results; 2) to immunize the segmentation model against the shifted-correlation effect, by removing the confounder that spuriously correlates objects in the background and the object-of-interests during training.

Learning causal features and removing confoundings usually require \textit{intervention}: \textit{fixing} the variable-of-interest while incorporating other variables in a fair way \cite{pearl2009causality}. This is similar to randomized controlled trials. To mitigate acquisition shift, a straightforward intervention is to train the model with images of a fixed cohort of patients that are taken under all possible acquisition processes. Since this is unrealistic, we resort to data-augmentation-based intervention \cite{ilse2020selecting} that incorporates possible acquisition shifts via simulation.

In this work, we propose a causality-inspired data augmentation approach for single-source domain generalization. It exposes the network to synthetic acquisition-shifted training samples that incorporate shifts in intensity/texture and shifted correlations.
Specifically, to efficiently synthesize diverse appearances (intensities and textures) without losing generality, we employ shallow convolutional networks with random weights that are sampled at each training iteration to augment images. As stable decision rules can hardly be formed on constantly-varying intensities/textures, the network would resort to domain-invariant features such as shapes. To remove the confounder that leads to the shifted-correlation effect, we first reveal that the image acquisition process naturally confounds objects in the background and the object-of-interests, in terms of their appearances. We then design a practical method for simulating and independently sampling the appearances of potentially confounded objects durining training. This is achieved by applying different appearance transformations in a spatially-variable manner, with the help of pseudo-correlation maps computed using unsupervised algorithms. The overall approach is used as additional stages following standard data augmentations. It is therefore generic to architectures of segmentation networks.
In summary, we make the following contributions:
\begin{itemize}
    \item We investigate single-source domain generalization problem for cross-domain medical image segmentation from a causal view. We propose a simple and effective causality-inspired data augmentation approach.
    \item We propose 1) global intensity non-linear augmentation (GIN) technique that efficiently transforms images to have diverse appearances via randomly-weighted shallow convolutional networks; 2) interventional pseudo-correlation augmentation (IPA) technique that removes the confounder that leads to the shifted-correlation effect. This is realized by independently resampling appearances of potentially confounded objects. These two components function as cores of the proposed approach.
    \item We build a comprehensive testing environment for single-source domain generalization for cross-domain medical image segmentation. It covers cross-modality, cross-sequence (MRI) and cross-center settings with various anatomical structures. We hope this testing environment to facilitate future works on domain robustness for medical image segmentation.
\end{itemize}

%% file: 2_related_works.tex
\section{Related works}
\subsection{Unsupervised domain adaptation and domain generalization for medical image segmentation}
Considerable efforts have been made to alleviate domain shift for deep networks. Unsupervised domain adaptation (UDA) transfers a model trained on a source domain to a target domain using unlabeled target-domain data \cite{ben2010theory}. Existing techniques are mainly based on distribution alignment \cite{tzeng2014deep,ganin2015unsupervised}, or self-training  \cite{zou2018unsupervised}. 
Multi-source domain generalization (MDG) usually learns domain-invariant features in an one-off manner from multiple source domains. Recent techniques include meta-learning\cite{dou2019domain,li2019episodic}, style transfer \cite{zhou2021domain,chartsias2017multimodal}, transfer learning\cite{carlucci2019domain}, dynamic networks \cite{chattopadhyay2020learning} and so on. In medical imaging, recent methods \cite{liu2020saml,liu2021semi} based on meta-learning \cite{finn2017model} have achieved promising results. 
However, for both UDA and MDG, target domain data or multi-source data is often unavailable due to privacy and/or cost concerns in medical settings. 

Single-source domain generalization requires training data from one domain only. 
Recent works such as \cite{huang2020self} propose to remove features that cause the largest loss gradients, which are believed to be domain-dependent. Liu et al. \cite{liu2021generalize} propose to unify statistics of image features, which control image styles \cite{huang2017arbitrary}, among images from different domains. 
A major stream of works exploits data augmentation: training the network on deliberately perturbed samples to improve network robustness to real-world perturbations \cite{krizhevsky2012imagenet,devries2017improved,zhang2017mixup,volpi2018generalizing}. We review these stream of methods later with more details. Some most recent works such as \cite{wang2021learning} employ multiple techniques: data augmentation \cite{xu2020robust}, adversarial training \cite{hjelm2018learning} and contrastive learning \cite{khosla2020supervised}.

\subsection{Data augmentation for domain robustness}
Theoretical analysis \cite{wu2020generalization} suggests that data augmentation improves generalization by enlarging the span of the data and by regularizing decision boundaries. In practice, Cutout \cite{devries2017improved} strengthens robustness against feature missing caused by domain shift, by partially occluding training images. Mixup \cite{zhang2017mixup,verma2019manifold} regularizes decision boundaries by interpolating among training samples. RandConv \cite{xu2020robust} drives the network to learn shape information, which is domain-invariant, by randomly altering image textures using linear filtering. Our method is closely related to RandConv \cite{xu2020robust}. However, we show in our experiments that the linear filtering in RandConv \cite{xu2020robust} is oversimplified for accounting for domain gaps that occur in real-world settings. Adversarial data augmentation generates image perturbations that easily flip predictions of classifiers \cite{miyato2018virtual,volpi2018generalizing,qiao2020learning,li2021progressive}.  

In medical imaging, Zhang et al. \cite{zhang2020generalizing} employ a stack of photometric and geometric transformations to training images to improve domain robustness.
Billot et al. \cite{billot2020learning} propose a contrast-agnostic brain MRI segmentation strategy, which synthesizes training examples by sampling from pre-built generative models of brain images. However, this method necessities well-defined generative models from segmentation labels to images, which is usually unavailable in most of medical imaging applications. Furthermore, these generative models are often oversimplified. AdvBias \cite{chen2020realistic} is specially designed for medical image segmentation. It employs an adversarial augmentation technique based on a multiplicative bias field model. It outperforms a series of competitive methods on cross-center MRI segmentation \cite{chen2021cooperative}.

\subsection{Leveraging causality for robust deep learning}
\label{subsubsec: relate_causal_invariance}
As discussed in Sec. \ref{sec: intro}, learning causalities and mitigating confoundings usually require \textit{causal intervention} \cite{pearl2009causality}. In causal intervention, the variable-of-interest in a causal relationship is \textit{fixed}, while other variables are fairly incorporated. A model then learns causalities from these intervened samples. For example, a model that recognizes a \texttt{camel} might be mistakenly focusing on the background: \texttt{desert} \cite{arjovsky2019invariant}, since most of pictures of camels are taken in deserts. In this case, intervention can be done by incorporating different backgrounds: training with pictures of camels that are taken in diverse backgrounds like grassland and city. By this mean the model would learns that it is the camel rather than the desert that \textit{cause} a \texttt{camel} label. Causal relationships are usually modeled using structural causal models (SCM)\cite{pearl2009causality,pearl1995causal}. The \textit{fixing} operation is usually noted as $do(\cdot)$ \cite{pearl2009causality}. The distribution $p(Y | do(X = \texttt{camel}))$, is called \textit{interventional distribution}. Compared with conditional distribution $p(Y | X = \texttt{camel})$ that reflects \textit{correlation} in the observed data, $p(Y | do(X = \texttt{camel}))$ reflects \textit{causation} \cite{pearl2009causality}.

Causal ideas have been used for discovering image features that are semantically essential and robust \cite{arjovsky2019invariant,mahajan2021domain,atzmon2020causal,mitrovic2020representation}. Invariant risk minimization \cite{arjovsky2019invariant} learns causal image representations by enforcing these representations to be Bayesian optimal in all environments. Mahajan et al. \cite{mahajan2021domain} improve domain robustness using contrastive losses. Atzmon et al. \cite{atzmon2020causal} propose a causal mechanism to generalize a model to novel samples with unseen combinations of attributions.

Our idea of using data-augmentation-based intervention is inspired by \cite{ilse2020selecting,mitrovic2020representation}. \cite{ilse2020selecting} proves that \textit{post-hoc} data augmentation theoretically commutes with "physical" intervention.
Mitrovic et al. \cite{mitrovic2020representation} derive a practical loss function for causality-based domain generalization. Different from \cite{mitrovic2020representation}, we focus on 
the unanswered practical problem of designing an augmentation model tailored to the real-world problem: cross-domain medical image segmentation. Our work is also related to causal weakly-supervised segmentation by Zhang et al. \cite{zhang2020causal}, as both works study the adverse effect of confoundings among objects on image segmentation. While Zhang et al. \cite{zhang2020causal} focus on the intra-domain scenario, we focus on the effect of confoundings under domain shift.

%% file: 3_method.tex
\section{Method}
\label{sec: method}
\input{figs/figures_tex/method_overall.tex}
Our causality-inspired data augmentation approach aims to improve network robustness against domain shift, in particular, shifts caused by the differences in acquisition processes \cite{glocker2019machine}. 
Based our analysis in Sec. \ref{sec: intro}, we propose to expose the network to training examples that incorporate simulated intensity/texture shifts and shifted correlations among objects. 

Specifically, our approach is a synergy of a \textit{global intensity non-linear augmentation} technique (GIN) and an \textit{interventional pseudo-correlation augmentation} technique (IPA). As shown in Fig. \ref{fig:method_overview}\textcolor{red}{-B}, GIN transfers training images to have diverse appearances while keeping the shapes of anatomical structures unchanged, discouraging the network from biasing towards appearances.
IPA resamples possible appearances of potentially spuriously correlated objects (due to confounding) in the background and the object-of-interests, in independent and diverse manners. This is implemented as spatially-variable blending between two GIN-augmented images. 
The entire approach functions as additional steps in a standard data augmentation pipeline.

In the following sections, we first introduce the general problem formulation of the proposed data augmentation approach. We introduce GIN in Sec. \ref{subsec: gin}, with a detailed reasoning behind its design-of-choices. IPA is introduced in Sec. \ref{subsec: ipa}, where we firstly reveal that it is the \textit{image acquisition process} that naturally confounds objects in the background and the object-of-interests. We then describe how IPA removes confoundings. Finally, we summarize the overall training process.

\subsection{Problem formulation}
\label{subsec: method_overview}
\noindent \textbf{A causal view of image generation and segmentation:} 
We first introduce the problem formulation of our data-augmentation-based single-source domain generalization approach. Inspired by recent works \cite{mitrovic2020representation,atzmon2020causal}, we model the data generation process and the (ideally, domain-invariant) segmentation process using the causal model shown in Fig. \ref{fig:method_overview}\textcolor{red}{-A}. Specifically, we make the following assumptions: 

\noindent 1) $A \rightarrow X \leftarrow C$ : An image $X$ is generated from two independent variables (factors): \textit{acquisition} $A$ and \textit{content} $C$. $C$ represents the shapes of underlying anatomical structures of the patient, while $A$ represents the acquisition process. The factor $A$ maps different types of tissues of the patient in the scanner into different pixel values in the image.

\noindent 2) $C \rightarrow S \rightarrow Y$: There exists an ideal domain-invariant representation $S$, determined by $C$. $S$ is in the form of feature maps of the deep layers of the network and it contains the shape information of the object-of-interests. The ground-truth segmentation mask $Y$ can be derived from $S$.  

\noindent 3) $X \rightarrow f_{\phi}(X) $: The segmentation network $f_{\phi}(\cdot)$ takes $X$ as an input and predicts $Y$, by implicitly estimating $S$. 

$A$ and $C$ are independent: changing $A$ does not affect $C$, $S$ and $Y$.
Of note, our discussion is constrained to acquisition shift. $C$ is assumed to be unchanged across the source domain and the target domain in our experiments. 

\noindent \textbf{Causal intervention for domain robustness:} According to \cite{mitrovic2020representation}, our argument that $S$ \textit{to be invariant to shifts of} $A$ can be formally written as follows:
\begin{align}
\small
\label{equ: inter_inv}
    p(Y | S, do(A=a_i)) = p(Y | S, do(A=a_j)), \forall a_i, a_j.
    \vspace{-5pt}
\end{align}
\normalsize
Here $p(Y | S, do(A=a_i))$ denotes the distribution raised by letting images to be generated from a specific acquisition process $A = a_i$ \cite{pearl2009causality,mitrovic2020representation}, for example MRI. Using a symmetric notation, we let $a_j$ to be another acquisition process, for example CT. Eq. \ref{equ: inter_inv} suggests that ideally, this distribution should remain the same regardless of the acquisition processes.

In practice, as shown in Fig.\ref{fig:method_overview}\textcolor{red}{-A}, we use a segmentation network $f_{\phi}(\cdot)$ parameterized by $\phi$ to predict $Y$. To make $f_{\phi}(\cdot)$ domain invariant, we implicitly estimate $S$ in the last layers of $f_{\phi}(\cdot)$. However, the condition in Eq. \ref{equ: inter_inv} cannot be directly used to train $f_{\phi}(\cdot)$, as ``physical'' interventions on $A$ (scanning patients under all possible acquisition processes) is impractical. Fortunately, Ilse et al. \cite{ilse2020selecting} have demonstrated that data augmentations can be used as ``virtual'' causal interventions. Therefore, we assume that for each $a_i$, there exists a photometric transformation function $\mathcal{T}_i(\cdot)$ being able to transform the image $X$ to be like from $a_i$ \footnote{Arguably this transformation can be interpreted as generating \textit{counterfactual} examples \cite{pearl2009causality,castro2020causality}: given an observed image from a certain acquisition process, asking how this image would look like, had it been generated from another imaging process. We do not label our data augmentation as counterfactual since the aim of our approach is not to synthesize the appearance of a \textit{specific} domain.}. We therefore have:
\begin{align}
\small
\label{equ: transform_func}
    p(Y| S, do(A=a_i)) \approx p(Y | f_{\phi}( \mathcal{T}_i (X) )).
\normalsize
\end{align}
Combining Eq. \ref{equ: inter_inv} and \ref{equ: transform_func} leads to a practical domain invariance condition: minimizing the difference between distributions raised by different photometric transformations $\mathcal{T}_i(\cdot)$ and $\mathcal{T}_j(\cdot)$ \cite{mitrovic2020representation}. By combining this domain invariance condition and the image segmentation loss, we can now derive our loss function (inspired by \cite{mitrovic2020representation}). For each iteration, we have
\begin{align}
    \label{equ: overll_obj}
    \small
    \mathcal{L}(\phi) & =  
    \mathcal{L}_{seg}( f_{\phi}(\mathcal{T}_i(\mbx)), \mby) + \mathcal{L}_{seg}( f_{\phi}(\mathcal{T}_j(\mbx)), \mby) \\ \nonumber +  & \lambda_{div} \mathcal{D}  (p(\mby | f_{\phi} (\mathcal{T}_i(\mbx)) ) \Vert p(\mby | f_{\phi} (\mathcal{T}_j(\mbx)) ) ).
    \normalsize
\end{align}
\normalsize
Here $(\mbx, \mby) \sim p(X,Y)$ denotes an image-label pair in the training dataset, $\mathcal{L}_{seg}(\cdot, \cdot)$ is a  segmentation loss, \textit{e.g.} cross-entropy; $\mathcal{T}_i(\cdot)$ and $\mathcal{T}_j(\cdot)$ are two different photometric transformations that simulates the effect of $a_i$ or $a_j$ respectively, randomly sampled from a family of photometric transformations at each iteration; $\mathcal{D}(\cdot \Vert \cdot) $ is the 
Kullback–Leibler divergence: measuring differences between two distributions; $\lambda_{div}$ is a weighting coefficient. Similar divergence terms have also been used for semi-supervised learning \cite{sajjadi2016regularization,miyato2018virtual}, although they are not derived from a causal perspective.

As implied by Eq. \ref{equ: overll_obj}, the photometric transformations $\{\mathcal{T}(\cdot)\}$ serve as the core of the domain invariance condition. Since the target-domain data is unavailable, we resort to build $\{ \mathcal{T}(\cdot) \}$ in a bottom-up manner, based on our analysis on ingredients of acquisition shift, as discussed in Sec. \ref{sec: intro}. We simulate $\{\mathcal{T}(\cdot)\}$ as a combination of GIN and IPA.

\subsection{Global intensity non-linear augmentation}
\label{subsec: gin}
\input{figs/figures_tex/gin_overview}
\noindent \textbf{Design-of-choices:} GIN is designed to efficiently transform image intensities and textures. We configure GIN as a family of piece-wise linear functions $g(\cdot) \in \mathcal{G}$, operating in pixel level or small local patch level in a spatially-invariant manner. These functions take a training image $\mbx$ from the source domain as input, and outputs an image with the same shape information but different intensities/textures, namely, $g(\cdot) : \mathbb{R}^{C_h \times H \times W} \rightarrow \mathbb{R}^{C_h \times H \times W}$, where $(H,W)$ to be the spatial size of a 2-D image $\mbx$, and $C_h$ to be the number of channels. As shown in Fig. \ref{fig: gin_overview}, transformations sampled from GIN are instantiated as shallow multi-layer convolutional networks $g_{\theta}(\cdot)$'s. These networks are composed with 1) random convolutional kernels $\theta$ sampled from Gaussian distributions $\mathcal{N}(0,I)$ with small receptive fields (to avoid over-blurring). 2) Leaky ReLU non-linearities between two neighboring convolutional layers (to make transformations non-linear). At each iteration, new $g_{\theta}(\cdot)$'s are sampled, yielding a variety of transformation functions. 
Inspired by \cite{xu2020robust}, we perform linear interpolation between the original and the output of the random network. 
In the end, as depicted in Fig. \ref{fig: gin_overview}, to constrain the energy of the augmented image, the output image is re-normalized to have the same Frobenius norm as the original input $\mbx$. We note the pure network part of the transformations $g_{\theta}(\cdot)$ as $g^{Net}_{\theta}(\cdot)$ (see Fig. \ref{fig: gin_overview}), and note a random interpolation coefficient sampled from uniform distribution $\mathcal{U}(0,1)$ as $\alpha$. We can write the transformed image $g_{\theta}(\mbx)$ as follows:
\begin{align}
    \small
    \label{equ: gin_output}
    g_{\theta}(\mbx) = \frac{\alpha  g^{Net}_{\theta}(\mbx) + (1-\alpha) \mbx }{ \Vert \alpha  g^{Net}_{\theta}(\mbx) + (1-\alpha) \mbx  \Vert}_{F} \cdot \Vert \mbx \Vert_{F},
    \normalsize
\end{align}
where $\Vert \cdot \Vert_{F}$ is the Frobenius norm.

By this mean, at each iteration, different intensities and textures are given to training images. As stable decision rules can hardly be built on randomly changing intensities/textures, the network would resort to invariant information like shapes. 

\noindent \textbf{Advantages:} We highlight major advantages of the above configurations: Firstly, GIN is based on generic assumptions on intensity/texture transformations. It therefore avoids being over-specific to a certain target domain(s). In addition, GIN is computationally efficient: it is in the form of shallow networks and therefore easily exploits GPUs for acceleration. Also, it is differentiable, and therefore can be be integrated into adversarial augmentation frameworks \cite{miyato2018virtual,volpi2018generalizing,qiao2020learning,chen2020realistic} to improved data efficiency.
\subsection{Interventional pseudo-correlation augmentation}
\label{subsec: ipa}
\label{subsubsec: spurious-correlation}
\input{figs/figures_tex/ipa_causal}
\noindent \textbf{Confounded objects in the background affect segmentation:} 
Recall in Sec. \ref{sec: intro}, spurious correlations (due to confoundings) between objects in the background and the object-of-interests in the source domain might be taken by a segmentation network as domain-specific clues for making predictions\footnote{In our preliminary experiment, we verified the existence of decision rules that are based on the background, in medical image segmentation: We distorted the backgrounds of images by randomly swapping patches of the background (therefore the global image statistics would remain unchanged.). We then tested a segmentation network with these background-distorted images, and have observed substantial performance downgrade compared with results on the original undistorted images. This phenomenon has also been verified in general computer vision and these spurious correlations are sometimes termed as \textit{context bias} \cite{hoyer2019grid,zhang2020causal}.}. These decision rules may break in the target domain, leading to performance downgrade. This is because confounded objects in the background that benefit segmentation in source domain, might not exist in the target domain. Alternatively, they might not co-shift in the same way as the object-of-interests. 

From the perspective of network architectures, objects in the background often affect predictions of the object-of-interests by the following ways: 1) Background features can affect global feature statistics at normalization layers \cite{burns2021limitations}, since feature statistics are usually calculated across all spatial locations. 2) The large receptive fields often make the pixels of the object-of-interests and those of the neighboring background to be inevitably perceived and processed together \cite{oktay2018attention}. 

\noindent \textbf{\noindent The acquisition process naturally confounds objects:} To mitigate the shifted-correlation effect, it is worthwhile to point out that it is the \textit{acquisition factor} $A$ that leads to confounding. It naturally creates spurious correlations between certain objects in the background and the object-of-interest\footnote{For the ease of illustration, in the following analysis, we focus on the scenario where only one object-of-interest is available. Our conclusion naturally holds for multi-class segmentation as well, and has been validated by our experiments on abdominal and cardiac segmentations.}. To demonstrate this, in an image $X$, we consider the patch of object-of-interest $X_f$ and the patch of a potentially correlated unlabeled object $X_b$ in the background. We zoom-in the causal relations in Fig. \ref{fig:method_overview}\textcolor{red}{-A} using $X_f$ and $X_b$, and redraw that in Fig. \ref{fig: ipa_causal_nov}\textcolor{red}{-A}. We can see: 

\noindent 1) $X_f \rightarrow f_{\phi}(X) \leftarrow X_b $: Although $X_f$ already contains sufficient information for delineating $Y$, in practice $X_b$ also affects the network features and the output, as both $X_f$ and $X_b$ are processed by $f_{\phi}(\cdot)$.

\noindent 2) $f_{\phi}(X) \leftarrow X_b \leftarrow A \rightarrow X_f$: More importantly, the confounding effect of $A$ that correlates $X_b$ and $X_f$ is revealed in the path $X_b \leftarrow A \rightarrow X_f$. This corresponds to the fact that given a certain acquisition process, the \textit{same} imaging physical mechanism that maps different tissues to different pixel values, applies to \textit{both} the object-of-interest and the objects in the background. Without such a path, appearances of $X_f$ and $X_b$ would vary independently in the training dataset. Stable correlations regarding their appearances could unlikely  be established and learned.

Of note, as we assume the content factor $C$ remain unchanged across domains, we ignore the confoundings caused by $C$, and omit $C$ in Fig. \ref{fig: ipa_causal_nov}. 

\noindent \textbf{Removing confounding by intervention:} We propose to mitigate the shifted correlation effect during the training stage, by removing the confounding $X_{b} \leftarrow A \rightarrow X_f$ using the intervention $do(X_f = \mbx_f)$ \cite{pearl2009causality}. This operation resamples the appearances of correlated objects $X_{b}$'s, independent of $X_f$. This intervention in effect removes $A \rightarrow X_f$. Formally, we are learning the interventional distribution $p( Y | do(X_f = \mbx_f) )$ based on the intervened causal diagram Fig. \ref{fig: ipa_causal_nov}\textcolor{red}{-B}: %
\begin{align}
    p(Y | do(X_f = \mbx_f) ) & = \sum_{\mbx_b} p (Y | \mbx_f, \mbx_b) p (\mbx_b) \nonumber \\
    & = \sum_{\mbx_b} \sum_{a}  p( Y | \mbx_f, \mbx_b) p(\mbx_b | a) p(a).
    \label{equ: ipa_backdoor}
\end{align}
Here $\mbx_f, \mbx_b \in \mbx; \  (\mbx, \mby) \sim p(X,Y); \ a \sim p(A)$ and $p(A) $ is a prior of possible acquisition processes. Eq. \ref{equ: ipa_backdoor} translates to independently sampling possible appearances of $X_b$.

Unfortunately, to compute Eq. \ref{equ: ipa_backdoor} we are faced with three practical issues: 1) 
we do not know which object is correlated with the object-of-interest and there is no ground-truth map of it; 2) there might be more than one objects in the background that correlates with the object-of-interest, and their effects might be entangled; 3) directly fixing $\mbx_f$ using ground-truth masks $\mby$ would make $\mbx_f$ unnaturally stand out from the background, providing shortcuts for the network to recognize $\mbx_f$. Also, this intervention \textit{cannot} be realized by GIN alone: GIN's transformation functions are spatially-invariant\footnote{If two objects share the same appearance, their appearances would remain the same after being transformed by GIN.}.
\input{figs/figures_tex/ipa_blender_nov}

\noindent \textbf{Spatially-variable blending:} As a practical solution, we employ \textit{interventional pseudo-correlation augmentation} (IPA), which approximates the intervention $do(X_f=\mbx_f)$. IPA is built on appearance transformations of GIN.
We use \textit{pseudo-correlation maps} as surrogates of label maps of $\mbx_b$'s, for allocating transformation functions to different pixels of the image: Pixels that correspond to different values in the pseudo-correlation map would be given different transformation functions. Pseudo-correlation maps are generated using the unsupervised algorithm \cite{de1978practical}. To account for different potential spurious correlations, we use different randomly-sampled maps at each iteration. To avoid the shortcuts caused by fixing $\mbx_f$ using $\mby$, we apply pseudo-correlation maps to both $\mbx_b$'s and $\mbx_f$ (\textit{i.e.} to the entire image).

To improve computation efficiency, as shown in Fig. \ref{fig: ipa_blender_nov}\textcolor{red}{-A}, we use pseudo-correlation maps as coefficients for blending pixels from two GIN-augmented versions of a same image. Considering a pseudo-correlation map $\mb{b} \in \mathbb{R}^{C_h \times H \times W}$ where all entries $b \in \mb{b}$ satisfy $b\in [0,1]$, we have the output image of IPA $\mathcal{T}_{1}(\mbx)$ as:
\begin{align}
\label{equ: ipa}
\small
    \mathcal{T}_{1}(\mbx; \theta_1, \theta_2, \mb{b}) = g_{\theta_1}(\mbx) \odot \mb{b} + g_{\theta_2}(\mbx) \odot (1-\mb{b}). 
    \normalsize
\end{align}
Here $\mathcal{T}_{1}(\cdot)$ denotes the overall combined effect of GIN and IPA, $\odot$ denotes the Hadamard product; $(g_{\theta_1}(\cdot), g_{\theta_2}(\cdot))$ are two random appearance transformations sampled from GIN. We simultaneously obtain one additional augmented image $\mathcal{T}_{2}(\mbx)$ by swapping the positions of $\mb{b}$ and $1-\mb{b}$ in Eq. \ref{equ: ipa}. Here the subscript $1$ or $2$ of $\mathcal{T}(\cdot)$'s denotes whether it is $g_{\theta_1}(\mbx)$ or $g_{\theta_2}(\mbx)$ to be multiplied with $\mb{b}$. Of note, this operation can also be interpreted as an extension to AugMix\cite{hendrycks2019augmix} which is designed for image classification. Different from AugMix, IPA necessities strict spatial correspondence between pixels and labels to ensure accuracy of this pixel-wise prediction. Also, the blending coefficients of IPA are intentionally made spatially variable to simulate our causal intervention.\\
\noindent \textbf{Pseudo-correlation maps:} As shown in Fig. \ref{fig: ipa_blender_nov}\textcolor{red}{-B}, we configure pseudo-correlation maps as a field of continuous randomly-valued scalars with low spatial frequency. They are interpolated from a lattice of randomly-valued control points, using cubic B-spline kernels \cite{de1978practical}, based on the efficient implementation from \cite{sandkuhler2018airlab}. Spacing between two neighboring control points is empirically set to be 1/4 of image length to avoid introducing unnaturally large image gradients. This configuration features the following advantages: 1) It allows spatially-variable intensity transformation while does not severely distort shape information due to its low spatial frequency. 2) It further increases diversities of appearances by interpolating between two appearances. 
\subsection{Training objective}
\label{subsec: training}
\input{algo1.tex}
\input{tables/dataset_details}

\input{tables/overall_perform}
The overall training is end-to-end using the loss function described in Eq. \ref{equ: overll_obj}. For the ease of implementation we let the output of $f_{\phi}(\cdot)$ to be in the form of raw logits, and let $p(\mby | f_{\phi}( \cdot) )$ to be the probabilities obtained by passing the output of $f_{\phi}(\cdot)$ to a softmax function. For $\mathcal{L}_{seg}$, we employ a sum of multi-class cross-entropy loss and soft Dice loss.
We set the weighting coefficient $\lambda_{div}$ in Eq. \ref{equ: overll_obj} to be 10.0, same as in \cite{xu2020robust}. After training, the segmentation network $f_{\phi}(\cdot)$ is ready to be directly applied to unseen testing domains. The overall algorithm flow is summarized in Algorithm \ref{alg: overall}.

%% file: figs/figures_tex/method_overall.tex
\definecolor{amber(sae/ece)}{rgb}{1.0, 0.49, 0.0}
\begin{figure*}[!htbp]
\centering
\includegraphics[width=0.85\textwidth]{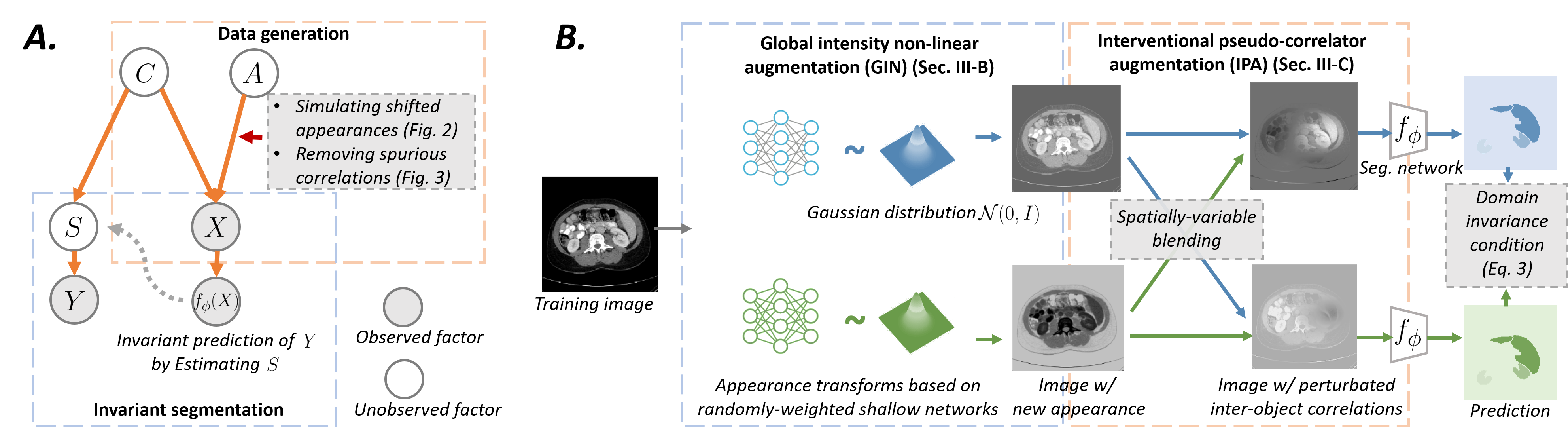}
\caption{\textbf{A. Data generation process of medical images:} An image $X$ is modeled as the effect of two independent factors: 1) a domain-independent content factor $C$ which represents the geometric shapes of the patient anatomy; 2) and a domain-dependent acquisition factor $A$ which controls image intensities and textures (appearances). \textbf{Mechanism of a domain-invariant segmentation network:} We assume that 1) there exists an ideal domain-invariant representation $S$ that contains shape information and is in the form of feature maps. $S$ is caused by $C$; 2) the pixel-wise ground-truth $Y$ can be derived from $S$. 
Our goal is to learn a network $f_{\phi}(\cdot)$ that predicts $Y$ by implicitly estimating $S$. To train $f_{\phi}(\cdot)$, our approach simulates different acquisition processes $A$'s and hence $X$'s. These simulated images encourage the network to distill $S$ from them.\\
\textbf{B. Workflow of the proposed data augmentation approach:} Our approach simulates different possible acquisition processes. GIN first transforms input images to have new appearances (intensities and textures), using randomly-weighted shallow networks. IPA then blends two GIN-augmented versions of a same image in a spatially-variable manner. This blending operation perturbs/randomizes spurious correlations among objects in an image, in terms of their appearances. Augmenting data with IPA allows the network to be immune to the \textit{shifted-correlation effect} (\textit{i.e.} the breaking of decision rules that are built on spurious correlations in the source domain, as described in Sec. \ref{sec: intro}). During training, the segmentation network $f_{\phi}(\cdot)$ is encouraged to make accurate and consistent predictions, unaffected by these simulated domain shifts. }
\label{fig:method_overview}
\end{figure*}

%% file: figs/figures_tex/gin_overview.tex
\begin{figure}[htb]
\centering
\includegraphics[width=0.9\linewidth]{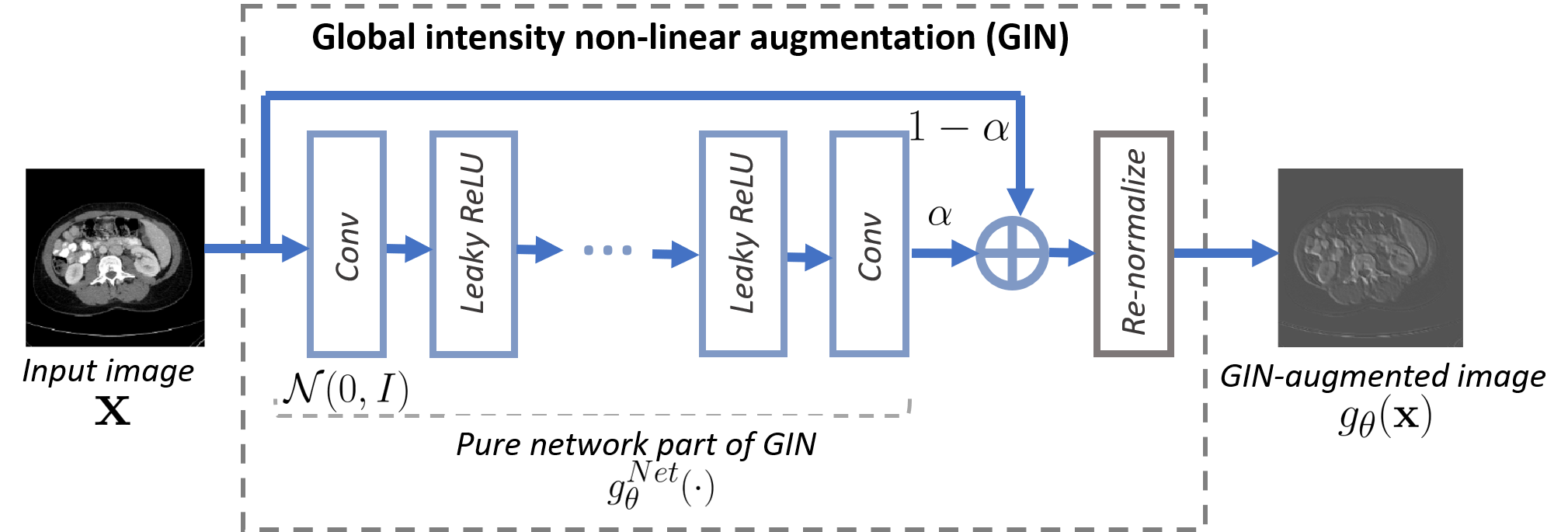}
\caption{\textbf{Illustration of the proposed global intensity non-linear augmentation (GIN) module.} It transforms image appearances using shallow convolutional networks with their weights randomly sampled at each iteration. It also contains Leaky ReLU's interleaved between convolutional layers. To maintain spatial resolutions of the input images, these random networks do not contain any downsampling operations.}
\label{fig: gin_overview}
\end{figure}

%% file: figs/figures_tex/ipa_causal.tex
\begin{figure}[htb]
\centering
\includegraphics[width=0.85\linewidth]{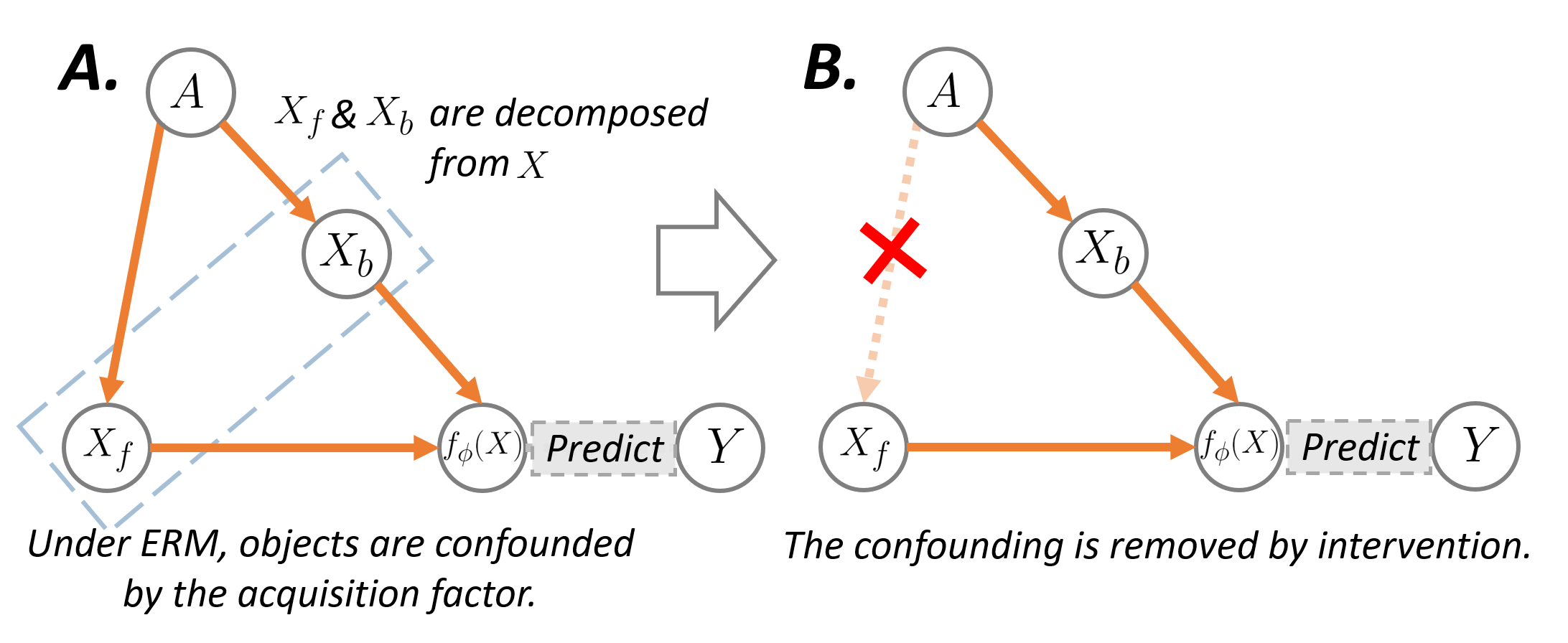}
\caption{\textbf{A. Causal graph illustrating spurious correlations between the object-of-interest $X_f$ and a potentially correlated object $X_b$}, where $X_f$ and $X_b$ are patches decomposed from a same image $X$. 
Spurious correlations between $X_f$ and $X_b$ are due to the acquisition factor $A$ which naturally confounds them.
In the source domain, the network $f_{\phi}(\cdot)$ may build domain-specific decision rules on these spurious correlations. In a shifted domain, these spurious correlations may break, and they are therefore detrimental to out-of-domain robustness of the network. This causal graph is a close-up of the data generation process in Fig. \ref{fig:method_overview}\textcolor{red}{-A}. The content factor $C$ in Fig. \ref{fig:method_overview}\textcolor{red}{-A} is assumed to be unchanged across domains and they are therefore omitted here. \\
\textbf{B. Removing the confounding caused by $A$}: This is achieved by the causal intervention $do(X_f = \mbx_f)$, which removes $A \rightarrow X_f$ in the causal graph. This intervention is conducted by independently sampling possible appearances of potentially correlated objects $X_b$'s.}
\label{fig: ipa_causal_nov}
\end{figure}

%% file: figs/figures_tex/ipa_blender_nov.tex
\begin{figure}[!t]
\centering
\includegraphics[width=0.9\linewidth]{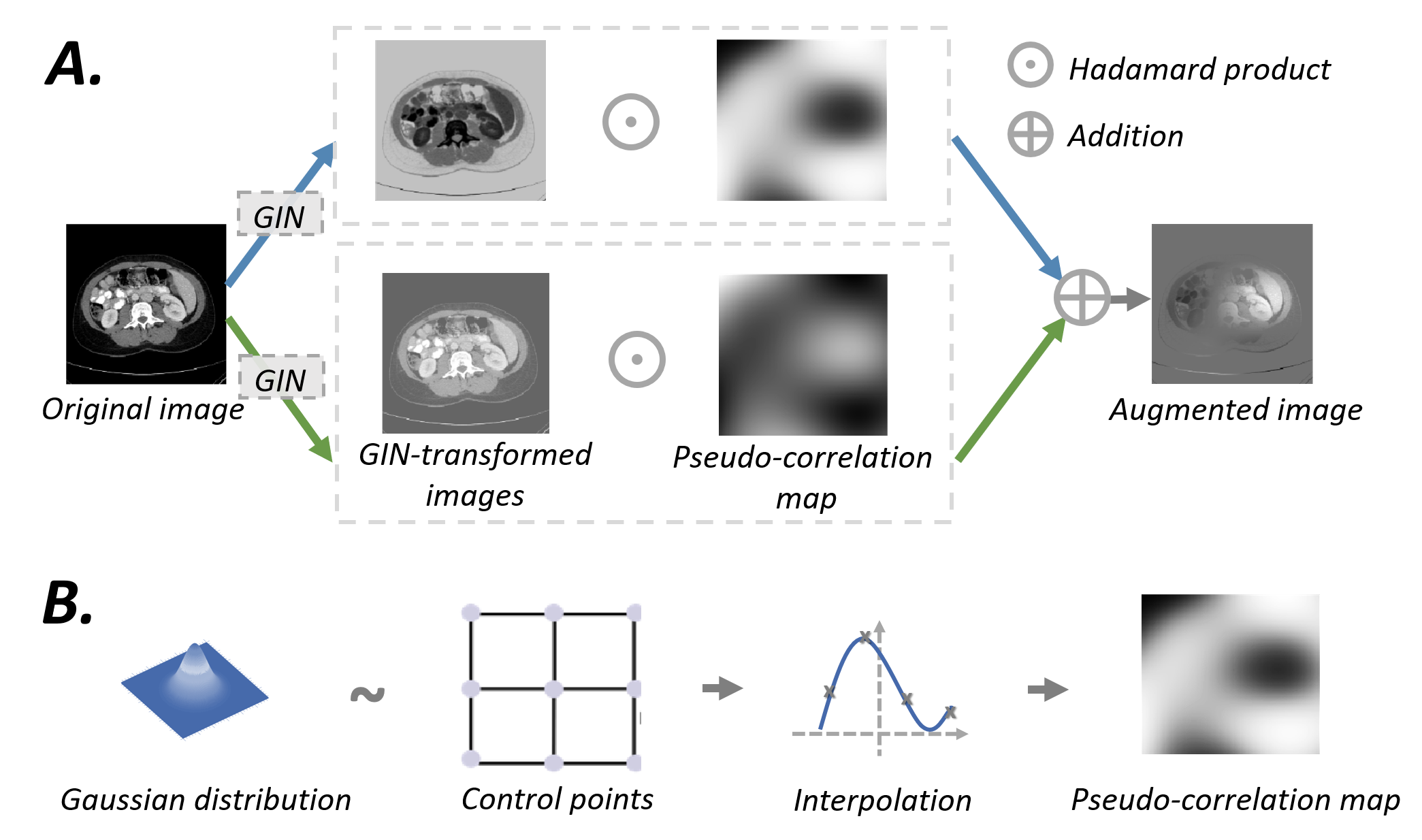}
\caption{\textbf{A. Implementation of IPA:}  In each iteration, a new pseudo-correlation map is generated, and it is used as pixel-wise coefficients for blending two GIN-transformed images. This is equivalent to assigning different appearance transformations to different pixels/patches, according to their corresponding values in the pseudo-correlation map. As the pseudo-correlation maps used in each iteration are different, during the training process, IPA can approximate the operation of resampling appearances of potentially confounded objects.\\
\textbf{B. Computing pseudo-correlation maps:} These maps are generated by interpolating along a lattice of randomly-valued control points.}
\label{fig: ipa_blender_nov}
\end{figure}

%% file: algo1.tex
\begin{algorithm}
\caption{End-to-end training with causality-inspired data augmentation}
  \begin{algorithmic}[1]
    \REQUIRE Training dataset $\{(\mbx, \mby )\}$ from the source domain, segmentation network $f_{\phi}(\cdot)$, global intensity non-linear augmentaton (GIN), interventional pseudo-correlation augmentation (IPA), number of iterations N, learning rate $l_r^{(t)}$ at iteration $t$.\\
    \FOR{ $t$ = 1 ... N}
    \STATE Sample an image-label pair ($\mbx^{(t)}$, $\mby^{(t)}$) from $\{(\mbx, \mby )\}$.
    \STATE Sample intensity/texture transformations $g_{\theta_1}^{(t)}(\cdot), g_{\theta_2}^{(t)}(\cdot)$ using GIN.
    \STATE Apply $g_{\theta_1}^{(t)}(\cdot), g_{\theta_2}^{(t)}(\cdot)$ as described in Eq. \ref{equ: gin_output} to $\mbx^{(t)}$ and obtain $g_{\theta_1}^{(t)}(\mbx^{(t)}), g_{\theta_2}^{(t)}(\mbx^{(t)})$.
    \STATE Compute a pseudo-correlation map $\mb{b}^{(t)}$ by interpolating along randomly valued control points as in Fig. \ref{fig: ipa_blender_nov}\textcolor{red}{-B}.
    \STATE Compute augmented images $\mathcal{T}_{1}^{(t)}(\mbx^{(t)}), \mathcal{T}_{2}^{(t)}(\mbx^{(t)})$ using IPA, as described in Eq. \ref{equ: ipa}. 
    \STATE Compute training loss $\mathcal{L}^{(t)}(\phi)$ using Eq. \ref{equ: overll_obj}.
    \STATE Update parameters of the segmentation network $f_{\phi}(\cdot)$: $\phi^{(t+1)} := \phi^{(t)} - l_r^{(t)} \frac{\partial \mcl{L}^{(t)}(\phi) } {\partial \phi}$.
    \ENDFOR 
  \end{algorithmic}
  \label{alg: overall}
\end{algorithm}

%% file: tables/dataset_details.tex
\begin{table*}[!ht]
    \centering
    \caption{Details of cross-domain segmentation datasets used in this study.}
    \setlength{\tabcolsep}{3pt}
    \resizebox{0.9\textwidth}{!}{
    \begin{tabular}{c|c|c|c|c|c|c }
    \toprule
    Name   & Label(s) & View & Split & Domain(s) & No. of 3-D scans & Origin \\ \hline
    \multirow{2}{*}{Abdominal CT-MRI} &  \multirow{2}{*}{\parbox{3cm}{ \centering Liver, L-kidney, R-kidney, Spleen }  } & 
    \multirow{2}{*}{Axial} & Source & Computed tomography (CT) & 30 &\cite{landman2015miccai}  \\
    & & & Target & T2 spectral presaturation with inversion recovery (SPIR) MRI & 20 &\cite{kavur2021chaos} \\ 
    \hline
    \multirow{2}{*}{Cardiac bSSFP-LGE} &  \multirow{2}{*}{\parbox{3cm}{\centering L-ventricle, Myocardium, R-ventricle}  } & 
    \multirow{2}{*}{Short-axis} & Source & Balanced steady-state free precession (bSSFP) MRI & 45 &\multirow{2}{*}{\cite{zhuang2020cardiac}}  \\
    & & & Target & Late gadolinium enhanced (LGE) MRI & 45 & \\ 
    \hline
    \multirow{2}{*}{Prostate Cross-center} &  \multirow{2}{*}{ Prostate }   & 
    \multirow{2}{*}{Axial} & 1 Source & \multirow{2}{*}{ Prostate MRI from 6 centers} & 30, 30, 19 &\cite{liu2020saml} \\
    & & & 5 Targets & & 13, 12, 12 & \cite{bloch2015nci,lemaitre2015computer,litjens2014evaluation} \\ 
\bottomrule
    \end{tabular}}
    
    \label{tab: dataset_detail}
\end{table*}

%% file: tables/overall_perform.tex
\begin{table*}[!ht]
    \centering
    \caption{Segmentation results on three cross-domain scenarios, where a model is trained on the source domain and tested on the target domain(s). Dice score is used as the evaluation metric. The highest scores are in \redc{red}, the second-highest scores are in \bluec{blue}.}
    \resizebox{0.9\textwidth}{!}{
    \begin{tabular}{c|cccc|c|ccc|c|c}
    \toprule
    \multirow{2}{*}{Method}   & \multicolumn{5}{c|}{Abdominal CT-MRI}  & \multicolumn{4}{c|}{Cardiac bSSFP-LGE} & Prostate Cross-center \\
    & Liver & R-Kidney & L-Kidney & Spleen & Average & L-ventricle & Myocardium & R-ventricle & Average & Prostate \\
    \hline
    Upper bound & 91.30 & 92.43 & 89.86 & 89.83 & 90.85 & 92.04 & 83.11 & 89.30 & 88.15 & 86.23 \\
    \hline
    ERM & 78.03 & 78.11& 78.45 & 74.65 & 77.31 & 86.06 & 66.98 & 74.94 & 75.99 & 56.59 \\
    Cutout \cite{devries2017improved} & \bluec{79.80} & \bluec{82.32} & 82.14 & 76.24 & 80.12 & 88.35 & 69.06 & 79.19 & 78.87 & \bluec{66.56}\\
    RSC \cite{huang2020self} & 76.40 & 75.79 & 76.60 & 67.56 & 74.09 & 87.06 & 69.77 & 75.69 & 77.51& 60.78\\
    MixStyle \cite{zhou2021domain}& 77.63 & 78.41 & 78.03 & 77.12 & 77.80 &85.78 & 64.23 & 75.61 & 75.21& 57.06\\
    AdvBias \cite{chen2020realistic} & 78.54 & 81.70 & 80.69 & 79.76 & 80.17 & 88.23 & 70.29 & 80.32 & 79.62& 61.98 \\
    RandConv \cite{xu2020robust} & 73.63 & 79.69 & \bluec{85.89} & \bluec{83.43} & \bluec{80.66} & \bluec{89.88} & \bluec{75.60} & \bluec{85.70} & \bluec{83.73} & 57.74 \\
    \hline
    Proposed & \redc{86.62} & \redc{87.48} &   \redc{
86.88} & \redc{84.27} & \redc{86.31} & \redc{90.35} & \redc{77.82} & \redc{86.87} & \redc{85.01} & \redc{70.37} \\
    \bottomrule
    \end{tabular}
    }
    \label{tab: performance}
\end{table*}

%% file: 4_experiments.tex
\section{Experiments}
\input{figs/figures_tex/qualitative_results.tex}
\subsection{Datasets and evaluation protocols}
The proposed approach is evaluated in three cross-domain settings: 1) cross-modality abdominal segmentation from CT to T2-SPIR MRI (Abdominal CT-MR), 2) cross-sequence cardiac segmentation from bSSFP MRI to LGE MRI (Cardiac bSSFP-LGE), and 3) prostate segmentation on MRI across six centers (Cross-center Prostate). Details of the datasets and the source-target splits are summarized in Table \ref{tab: dataset_detail}.
All datasets are originally in 3-D and have been reformatted to 2-D, then resized to 192$\times$192, and padded along the channel dimension to fit into the network. For the abdominal CT dataset, we applied a window of [-275, 125] \cite{ouyang2020self} in Housefield values. For all MRI images, we clipped the top 0.5\% of the histograms. We normalized all the 3-D scans to have zero mean and unit variance. For fairness of comparisons, for all the methods evaluated (including ERM), conventional geometric augmentations: affine transformations and elastic transformations; and intensity augmentations: brightness, contrast, gamma transformations and additive Gaussian noises were applied by default.

We employed the commonly-used Dice score (0-100) as the evaluation metric for measuring the overlap between the prediction and the ground truth. For abdominal and prostate segmentations, for the source domain, we used a 70\%-10\%-20\% split for training, validation and testing sets; for the target domain(s), we used all the images for testing, same as in \cite{liu2020saml}. For cross-center prostate segmentation, each time we took one domain as the source and the rest five domains as targets, and we computed Dice scores averaged by target domains.
This 1-versus-5 experiment is repeated for each of all six domains. For cardiac segmentation, we employed the same data split as in the cross-sequence segmentation challenge \cite{zhuang2020cardiac}.

\subsection{Network architecture and training configurations}
We configured the segmentation network $f_{\phi}(\cdot)$ as a U-Net \cite{ronneberger2015u}, the most commonly used network architecture for medical image segmentation, with an EfficientNet-b2 backbone \cite{tan2019efficientnet}. For our proposed method, we trained the segmentation network using an Adam optimizer \cite{kingma2014adam} with an initial learning rate of $3\times10^{-4}$ with learning rate decay. We evaluated our method at the 2k-th epoch where the learning rate decays to zero.

\subsection{Quantitative and qualitative results}
We compared our method with the empirical risk minimization (ERM) baseline and several recent single-source domain generalization methods. Among them Cutout \cite{devries2017improved} enforces the model to be robust to corruptions by deliberately removing patches from training images. RSC \cite{huang2020self} defines features that lead to the largest gradients as non-robust features and removes them in training. MixStyle \cite{zhou2021domain} synthesize novel domains by mixing instance-level feature statistics \cite{huang2017arbitrary}. AdvBias \cite{chen2020realistic} is designed for medical images. It augments images using adversarial perturbations. Closely related to our work is RandConv \cite{xu2020robust}, which employs a random linear intensity transformation model to synthesize novel domains. 

Table \ref{tab: performance} summarizes performances on three cross-domain segmentation scenarios, where a network is trained on the source domain and evaluated on the target domain(s). The proposed approach consistently outperforms peer methods. In particular, the performance gains of our method compared with the closely-related RandConv \cite{xu2020robust} suggest that our approach simulates domain shifts in a more effective manner, leading to stronger robustness upon unseen domains. We also provide the upper bounds: \textit{i.e. }training and testing in the target domain, in Table \ref{tab: performance}.  Qualitative examples are shown in Fig. \ref{fig: qualitative}.

To visualize the feature spaces, in Fig. \ref{fig: tsne} we show t-SNE of the target domain features collected at the last hidden layer of abdominal segmentation networks. As can be seen, for our proposed method, for the same class, target domain features stay close to those of the source domain; while features of different classes are separated.
\input{figs/figures_tex/tsne.tex}
\subsection{Ablation studies}
\subsubsection{Configurations of GIN}
\input{tables/ablation_gin}
To investigate the effect of configurations of GIN on generalization performance, we conducted ablation studies on two key design-of-choices: the number of convolutional layers and the number of channels in hidden layers. Intuitively, only one or two layers may be insufficient for simulating non-linear transformations across domains in real world, while a too-large number of layers may lead to unrealistically aggressive augmentations that deviate from reality. 
The effect of numbers of channels in hidden layers is difficult to conjecture, due to the non-linearity of GIN.

We show quantitative results in Table \ref{tbl: abl_gin} by varying the number of layers, and the number of channels in hidden layers from the default setting (4 layers and 2 channels). These experiments were conducted under the abdominal segmentation scenario, with IPA turned off for the ease of analysis. The left column of Table \ref{tbl: abl_gin} agrees with our intuition regarding the number of convolutional layers.
\input{tables/ablation_bias_type}
\subsubsection{Configurations of IPA}
\label{subsubsec: mask_compare}
To examine the effect of interventional pseudo-correlation augmentation, we performed ablation study by removing IPA from the proposed approach. The results in the first two rows of Table \ref{tab: type_mask} validate the benefit of mitigating the shifted-correlation effect using IPA, especially for the cardiac and the prostate settings, where $p$-values $< 0.001$ under Wilcoxon signed-rank tests.
\input{figs/figures_tex/superpix_map}

As a further exploration, we also examined an alternative design of pseudo-correlation maps: superpixels that are randomly sampled at each iteration \cite{felzenszwalb2004efficient}, as depicted in Fig. \ref{fig: superpix_map}. We present its results in the last row of Table \ref{tab: type_mask}. We conjecture the sub-optimal performance of the superpixel-based maps is due to the fact that superpixels often coincides with real ground-truth masks $\mby$'s, making $\mbx_f$'s unnaturally stand out and thus become shortcuts for the network.

%% file: figs/figures_tex/qualitative_results.tex
\begin{figure*}[htb]
\centering
\includegraphics[width=0.88\linewidth]{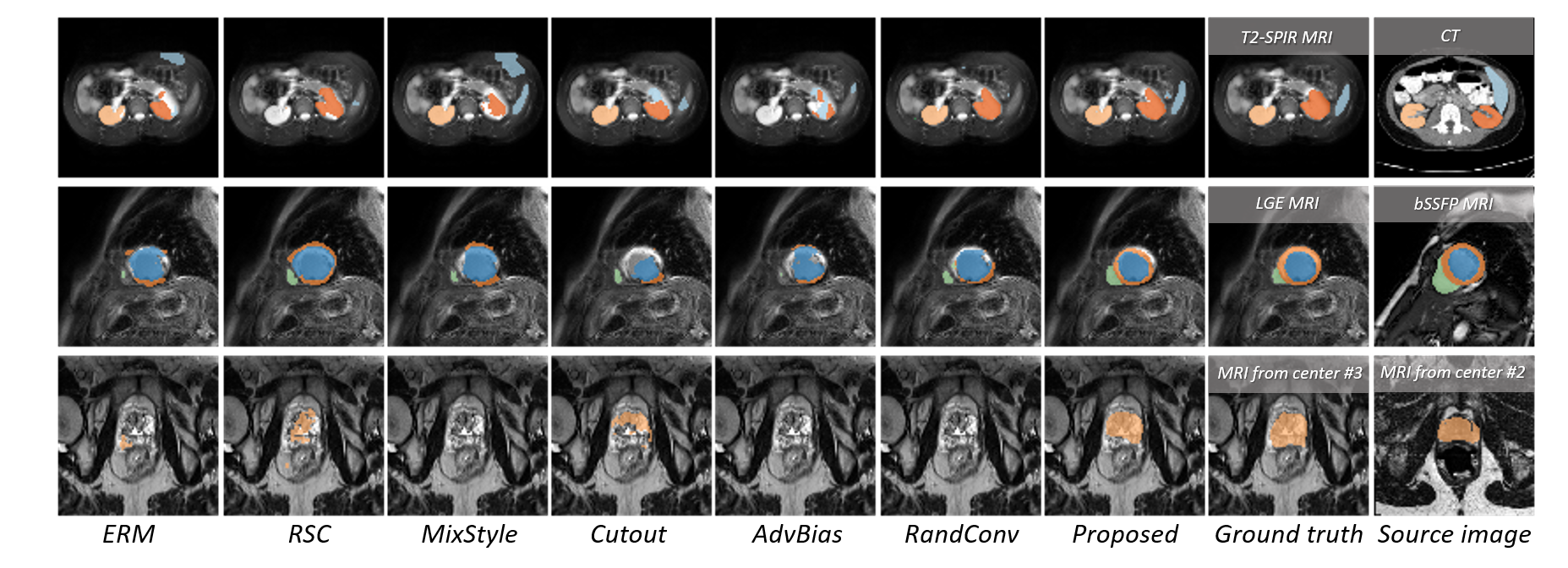}
\caption{ Qualitative results on cross-domain segmentation under three scenarios: Abdominal CT-MRI (top row), Cardiac bSSFP-LGE (middle row) and Prostate Cross-center (bottom row). Examples of source domain (training dataset) images are shown in the rightmost column.}
\label{fig: qualitative}
\end{figure*}

%% file: figs/figures_tex/tsne.tex
\begin{figure}[h]
\centering
\includegraphics[width=0.85\linewidth]{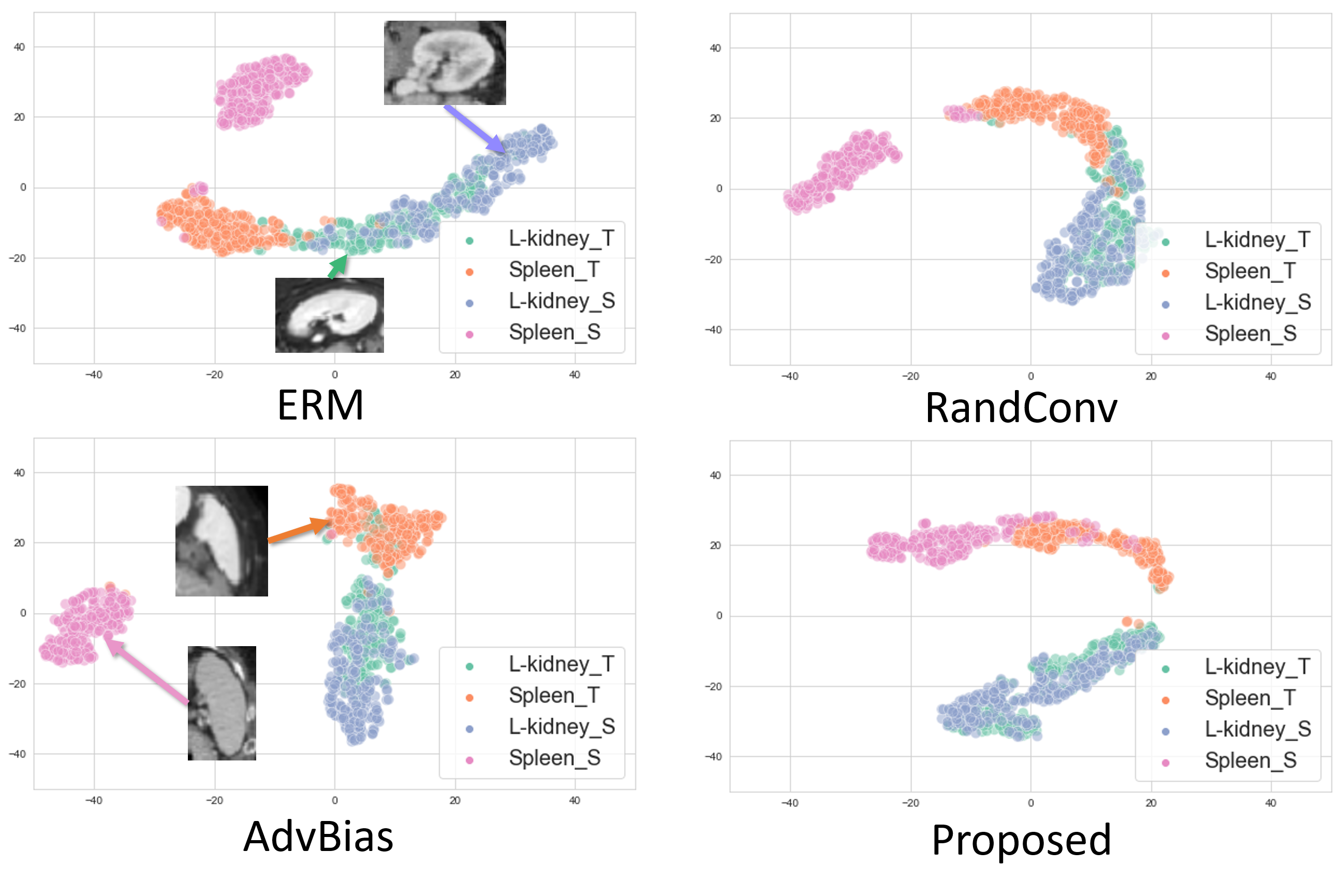}
\caption{t-SNE visualizations of the last hidden layer features of segmentation networks, trained using different domain generalization techniques, for the Abdominal CT-MR segmentation scenario. Suffixes \_S or \_T denote the source (CT) or the target domain (MRI).
}
\label{fig: tsne}
\end{figure}

%% file: tables/ablation_gin.tex
\begin{table}[]
    \centering
    \caption{Effects of number of total convolutional layers (left) and number of channels in hidden layers (right) on the performance of Abdominal CT-MRI segmentation.}
    \setlength{\tabcolsep}{3pt}
    \resizebox{0.7\linewidth}{!}{
\begin{tabular}{cc|cc}
\toprule
No. of layers & \multicolumn{1}{c|}{\begin{tabular}[c]{@{}c@{}}Average\\ Dice Score\end{tabular}} & \multicolumn{1}{c}{\begin{tabular}[c]{@{}c@{}}No. of channels\\ in hidden layers\end{tabular}} & \multicolumn{1}{c}{\begin{tabular}[c]{@{}c@{}}Average \\ Dice Score\end{tabular}} \\
\hline
2 & 84.26 & 2 (reported) & 86.04 \\
4 (reported) & 86.04 & 4 & 85.49 \\
8 & 85.66 & 8 & 83.21 \\
16 & 79.17 & 16 & 81.90 \\
    \bottomrule
    \end{tabular}
    }
\label{tbl: abl_gin}
\end{table}

%% file: tables/ablation_bias_type.tex
\begin{table*}[h]
    \centering
    \caption{Ablation study on interventional pseudo-correlation augmentation. The highest scores are in \redc{red}, the second-highest scores are in \bluec{blue}.}
    \setlength{\tabcolsep}{3pt}
    \resizebox{0.9\textwidth}{!}{
    \begin{tabular}{c|cccc|c|ccc|c|c}
    \toprule
    \multirow{2}{*}{Configuration}   & \multicolumn{5}{c|}{Abdominal CT-MRI}  & \multicolumn{4}{c|}{Cardiac bSSFP-LGE} & \multicolumn{1}{c}{Prostate Cross-center} \\
    & Liver & R-Kidney & L-Kidney & Spleen & Average & L-ventricle & Myocardium & R-ventricle & Average & Prostate \\
    \hline
    GIN-only & \redc{87.34} & \bluec{86.40} & \redc{87.18} & 83.25& \bluec{86.04} & 90.04 & 75.74 & 85.71 & 83.83 & 65.10 \\ 
    \hline
    GIN + IPA (reported)& 86.62 & \redc{87.48} &   \bluec{86.88} & \redc{84.27} & \redc{86.31} & \bluec{90.35}& \bluec{77.82} & \bluec{86.87} & \bluec{85.01} & \redc{70.37}\\
    GIN + IPA (superpixel-based) & \bluec{87.26} & 85.66 & 85.14 & 82.30 & 85.09 & \redc{91.07} & \redc{79.63} & \redc{87.19} & \redc{85.97} & \bluec{66.53}\\
    \bottomrule
    \end{tabular}
    }
    \label{tab: type_mask}
\end{table*}

%% file: figs/figures_tex/superpix_map.tex
\begin{figure}[htb]
\centering
\includegraphics[width=0.95\linewidth]{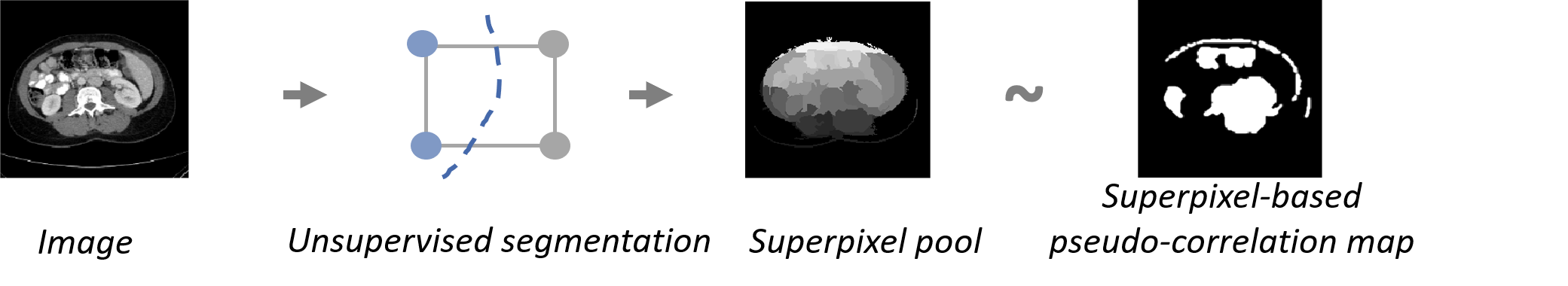}
\caption{An alternative configuration of pseudo-correlation maps: randomly sampled superpixels.}
\label{fig: superpix_map}
\end{figure}

%% file: 5_discussion.tex
\section{Discussion and conclusion}
Domain robustness has been a challenge for deep learning based medical image computing for a long time. In this work, we propose a causality-inspired data augmentation approach for single-source domain generalization. From a methodological perspective, while previous multi-source domain generalization (MDG) and unsupervised domain adapation (UDA) methods are \textit{top-down} solutions that learn \textit{a priori} knowledge of out-of-domain data (assumed to be available), our data augmentation is a \textit{bottom-up} approach based on the causal mechanism of acquisition shifts. Although challenging, pursing causalities and designing bottom-up methods encourage further theoretical investigations on domain shift, which in turn facilitates more principled techniques for robust learning. From a practical perspective, unlike UDA or MDG, our method does not require target-domain data or multi-source data to be available during training. Also, compared with UDA, our method is easier to deploy in real world: it does not require fine-tuning on the target domain (which more or less relies on expertise). Compared to peer single-source generalization techniques, our approach demonstrates consistently superior performances in our experiments. 

In the current approach several limitations remain: First, although consistent performance gains are shown in all three scenarios, some crucial hyper-parameters like the number of layers in GIN and the configurations of IPA still require empirical choices. A more elegant augmentation technique that requires less empirical choices is desirable. 
In addition, as domain-specific information is suppressed during training, our method experiences slight performance downgrades in source-domain testing sets on abdominal CT ($\sim$ -2.0/100) and prostate MRI's ($\sim$ -1.0/100), compared to the ERM counterparts. Potential solutions might reside in network architecture side, for example, a dynamic architecture which adaptively balances domain-specific and domain-invariant features.

Starting from the current methodology, several potential extensions arise: For example, as our method can efficiently produce huge amount of domain-shifted images, it is natural to combine the proposed method with multi-source domain generalization techniques \cite{dou2019domain,liu2020saml}. In addition, as our work focuses on image appearance, it is interesting to design methods targeting at domain shifts in terms of anatomical shapes. 